\documentclass[12pt]{article}
\RequirePackage{amsthm,amsmath,amsbsy,amsfonts}
\usepackage{graphicx,psfrag,epsf}
\usepackage{enumerate}
\usepackage{natbib}

\newcommand{\blind}{0}

\numberwithin{equation}{section}
\theoremstyle{plain}

\usepackage{scalerel}
\usepackage{amssymb}
\usepackage[usenames]{color}

\usepackage{ifpdf}

\usepackage{graphicx}
\ifpdf
\usepackage{epstopdf}
\DeclareGraphicsExtensions{.eps,.pdf}
\else
\usepackage{epsfig}
\fi

\usepackage{subfig}

\ifpdf
\usepackage[colorlinks,urlcolor=blue,citecolor=blue,linkcolor=blue]{hyperref}
\else
\newcommand{\href}[2]{{#2}}
\usepackage{url}
\fi

\ifpdf
\newcommand{\Sec}[1]{\hyperref[sec:#1]{Section~\ref*{sec:#1}}} 
\newcommand{\App}[1]{\hyperref[sec:#1]{Appendix~\ref*{sec:#1}}} 
\newcommand{\Eqn}[1]{\hyperref[eq:#1]{{\rm (\ref*{eq:#1})}}} 
\newcommand{\Part}[1]{\hyperref[part:#1]{(\ref*{part:#1})}} 
\newcommand{\Fig}[1]{\hyperref[fig:#1]{Figure~\ref*{fig:#1}}} 
\newcommand{\Tab}[1]{\hyperref[tab:#1]{Table~\ref*{tab:#1}}} 
\newcommand{\Thm}[1]{\hyperref[thm:#1]{Theorem~\ref*{thm:#1}}} 
\newcommand{\Lem}[1]{\hyperref[lem:#1]{Lemma~\ref*{lem:#1}}} 
\newcommand{\Prop}[1]{\hyperref[prop:#1]{Proposition~\ref*{prop:#1}}} 
\newcommand{\Cor}[1]{\hyperref[cor:#1]{Corollary~\ref*{cor:#1}}} 
\newcommand{\Def}[1]{\hyperref[def:#1]{Definition~\ref*{def:#1}}} 
\newcommand{\Alg}[1]{\hyperref[alg:#1]{Algorithm~\ref*{alg:#1}}} 
\newcommand{\Ex}[1]{\hyperref[ex:#1]{Example~\ref*{ex:#1}}} 
\newcommand{\As}[1]{\hyperref[as:#1]{Assumption~{\rm\ref*{as:#1}}}} 
\newcommand{\Reg}[1]{\hyperref[as:#1]{Condition~\ref*{reg:#1}}} 
\newcommand{\AlgLine}[2]{\hyperref[alg:#1]{line~\ref*{line:#2} of Algorithm~\ref*{alg:#1}}}
\newcommand{\AlgLines}[3]{\hyperref[alg:#1]{lines~\ref*{line:#2}--\ref*{line:#3} of Algorithm~\ref*{alg:#1}}}
\else
\newcommand{\Sec}[1]{{Section~\ref{sec:#1}}} 
\newcommand{\App}[1]{{Appendix~\ref{sec:#1}}} 
\newcommand{\Eqn}[1]{{(\ref{eq:#1})}} 
\newcommand{\Part}[1]{{(\ref{part:#1})}} 
\newcommand{\Fig}[1]{{Figure~\ref{fig:#1}}} 
\newcommand{\Tab}[1]{{Table~\ref{tab:#1}}} 
\newcommand{\Thm}[1]{{Theorem~\ref{thm:#1}}} 
\newcommand{\Lem}[1]{{Lemma~\ref{lem:#1}}} 
\newcommand{\Prop}[1]{{Property~\ref{prop:#1}}} 
\newcommand{\Cor}[1]{{Corollary~\ref{cor:#1}}} 
\newcommand{\Def}[1]{{Definition~\ref{def:#1}}} 
\newcommand{\Alg}[1]{{Algorithm~\ref{alg:#1}}} 
\newcommand{\Ex}[1]{{Example~\ref{ex:#1}}} 
\newcommand{\As}[1]{{Assumption~\ref{as:#1}}} 
\newcommand{\Reg}[1]{{R~\ref{reg:#1}}} 
\newcommand{\AlgLine}[2]{{line~\ref{line:#2} of Algorithm~\ref{alg:#1}}}
\newcommand{\AlgLines}[3]{{lines~\ref{line:#2}--\ref{line:#3} of Algorithm~\ref{alg:#1}}}
\fi

\usepackage{tikz}
\usetikzlibrary{arrows}
\newtheorem{theorem}{Theorem}

\newtheorem*{corollary}{Corollary}

\newtheorem*{lemma}{Lemma}

\theoremstyle{definition}

\theoremstyle{remark}

\newcommand{\Real}{\mathbb{R}}
\newcommand{\diam}{\operatorname{diam}}
\newcommand{\conv}{\operatorname{conv}}

\newcommand{\amp}{\mathop{\:\:\,}\nolimits}

\begin{document}

\def\spacingset#1{\renewcommand{\baselinestretch}%
{#1}\small\normalsize} \spacingset{1}

\if0\blind
{
  \title{\bf Recovering Trees with Convex Clustering}
  \author{Eric C. Chi\thanks{
    Department of Statistics, North Carolina State University, Raleigh, NC 27695-8203 (E-mail: eric$\_$chi@ncsu.edu)} \, 
    and
    Stefan Steinerberger\thanks{Department of Mathematics, Yale University, New Haven, CT,  06520-8283 (E-mail: stefan.steinerberger@yale.edu).}    \\}
    \date{}
  \maketitle
} \fi

\if1\blind
{
  \bigskip
  \bigskip
  \bigskip
  \begin{center}
    {\LARGE\bf Title}
\end{center}
  \medskip
} \fi

\bigskip
\begin{abstract}
Convex clustering refers, for given $\left\{x_1, \dots, x_n\right\} \subset \Real^p$, to the minimization of 
\begin{eqnarray*}
u(\gamma) & = & \underset{u_1, \dots, u_n }{\arg\min}\;\sum_{i=1}^{n}{\lVert x_i - u_i \rVert^2} + \gamma \sum_{i,j=1}^{n}{w_{ij} \lVert u_i - u_j\rVert},\\
\end{eqnarray*}
where $w_{ij} \geq 0$ is an affinity that quantifies the similarity between $x_i$ and $x_j$. We prove that if the affinities $w_{ij}$ reflect a tree structure in the $\left\{x_1, \dots, x_n\right\}$, then the convex clustering solution path reconstructs the tree exactly. The main technical ingredient implies the following combinatorial byproduct:  for every set $\left\{x_1, \dots, x_n \right\} \subset \Real^p$ of $n \geq 2$ distinct points,
there exist at least $n/6$ points with the property that for any of these points $x$ there is a unit vector $v \in \Real^p$ such that, when viewed from $x$, `most' points lie in the direction $v$
\begin{eqnarray*}
\frac{1}{n-1}\sum_{i=1 \atop x_i \neq x}^{n}{ \left\langle \frac{x_i - x}{\lVert x_i - x \rVert}, v \right\rangle} & \geq & \frac{1}{4}.
\end{eqnarray*}
\end{abstract}

\noindent%
\vfill

\newpage
\spacingset{1.45} 

\section{Introduction}
Hierarchical clustering is a fundamental unsupervised learning task, whose aim is to organize a collection of points into a tree of nested clusters. To reinforce the idea that we seek a collection of nested clusters, we will often also refer to clusters as folders in this paper. 

\begin{figure}[t]
\centering
\includegraphics[scale = 0.4]{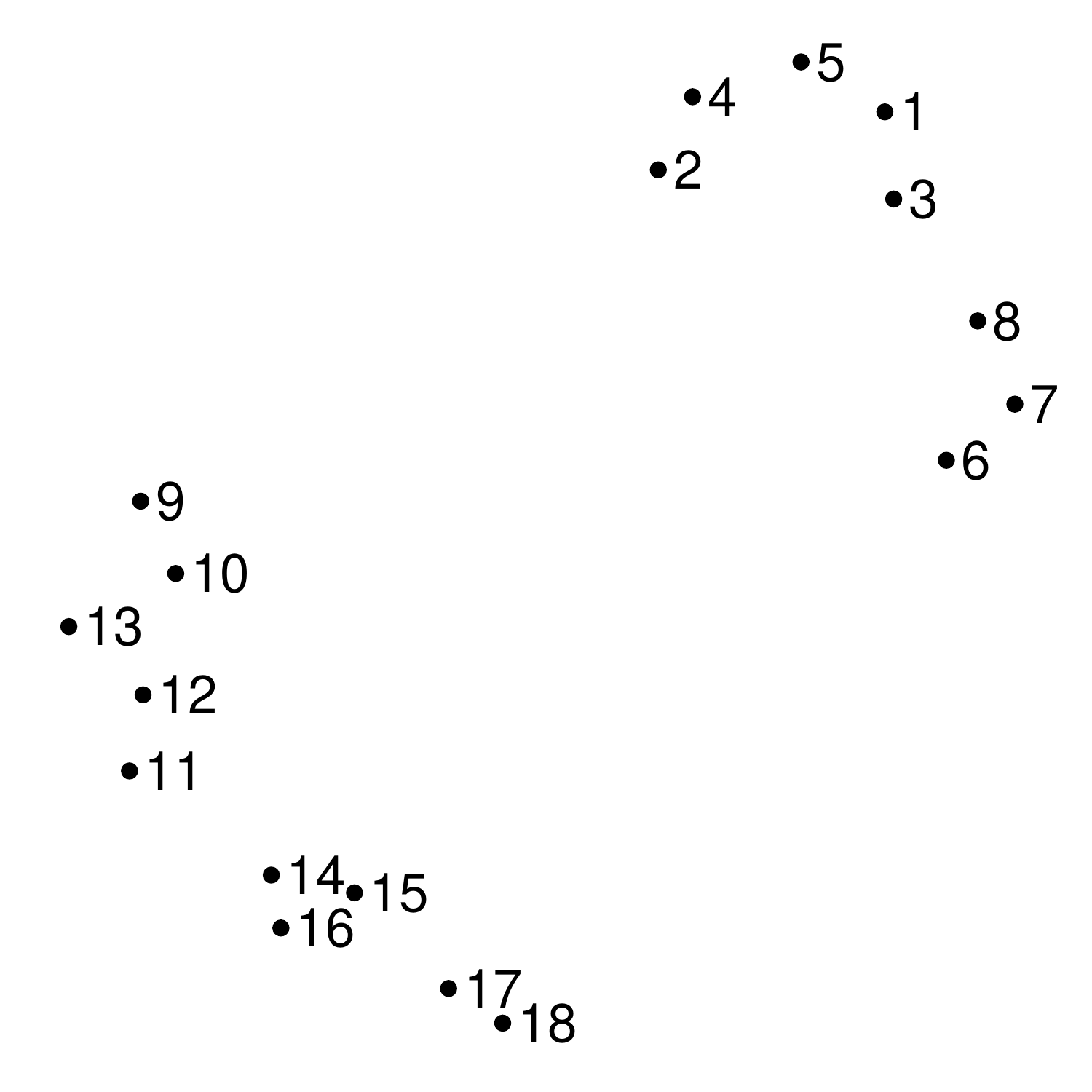}
\caption{Eighteen points in $\Real^2$ to organize. \label{fig:scatter}}
\end{figure}

As an illustration, 
\Fig{scatter} shows a collection of points in $\Real^2$, labeled 1 to 18,  that we seek to organize.  
Based on the Euclidean distances between the points, an intuitive organization is the following hierarchy of nested clusters. At the finest and first level of clustering, we partition the set $\{1, \ldots, 18\}$ into five subsets or folders:
\begin{eqnarray*}
F_{1,1} \amp = \amp \{1, 2, 3, 4, 5\},\; F_{1,2} & = & \{6, 7, 8\},\; F_{1,3} \amp = \amp \{9, 10, 11, 12, 13\},\\
F_{1,4} \amp = \amp \{14, 15, 16\}, &\text{and}& F_{1,5} \amp = \amp \{17, 18\}.
\end{eqnarray*}
At the second level of clustering, we merge the folders from the first level into a partition of two folders: $F_{2,1} = F_{1,1} \cup F_{1,2}$ and $F_{2,2} = F_{1,3} \cup F_{1,4} \cup F_{1,5}.$

\begin{figure}[h]
\centering
\includegraphics[scale = 0.343]{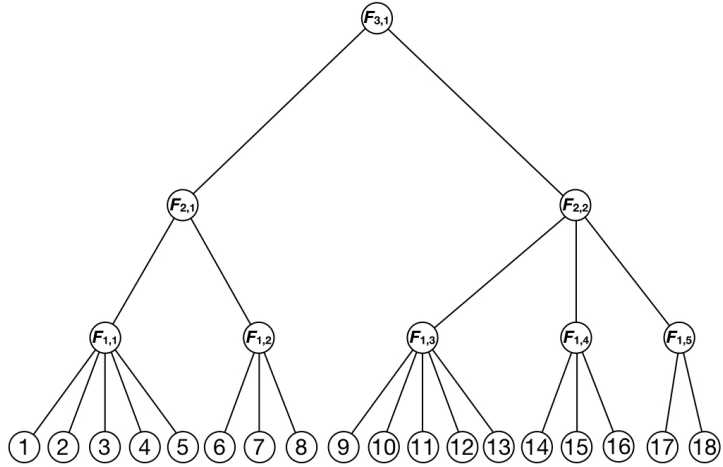}
	\caption{Partition Tree.\label{fig:partition_tree}}
\end{figure}

Finally, at the third level of clustering, we merge the folders from the second level into a single folder: $F_{3,1} = F_{2,1} \cup F_{2,2}.$ \Fig{partition_tree} illustrates the described tree organization. Since each level of the tree consists of a partition of the data points, we refer to such hierarchical organizations as ``partition trees."

\begin{figure}[t]
    \centering
	\begin{tabular}{cc}
	     \subfloat[Single-linkage tree]{\includegraphics[scale = 0.45]{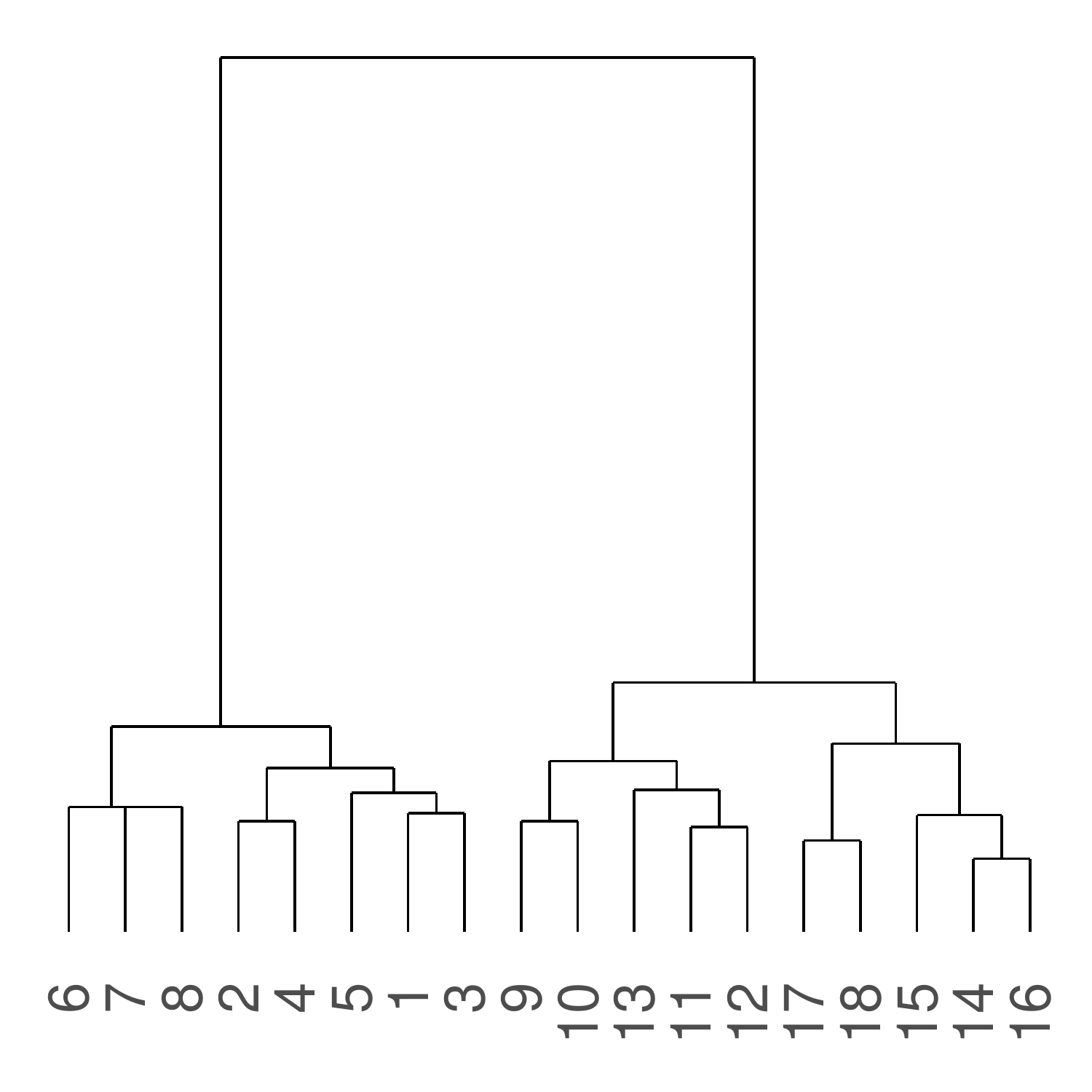} 
		\label{fig:hclust_single}} 		
	    & \subfloat[Average-linkage tree]{\includegraphics[scale = 0.45]{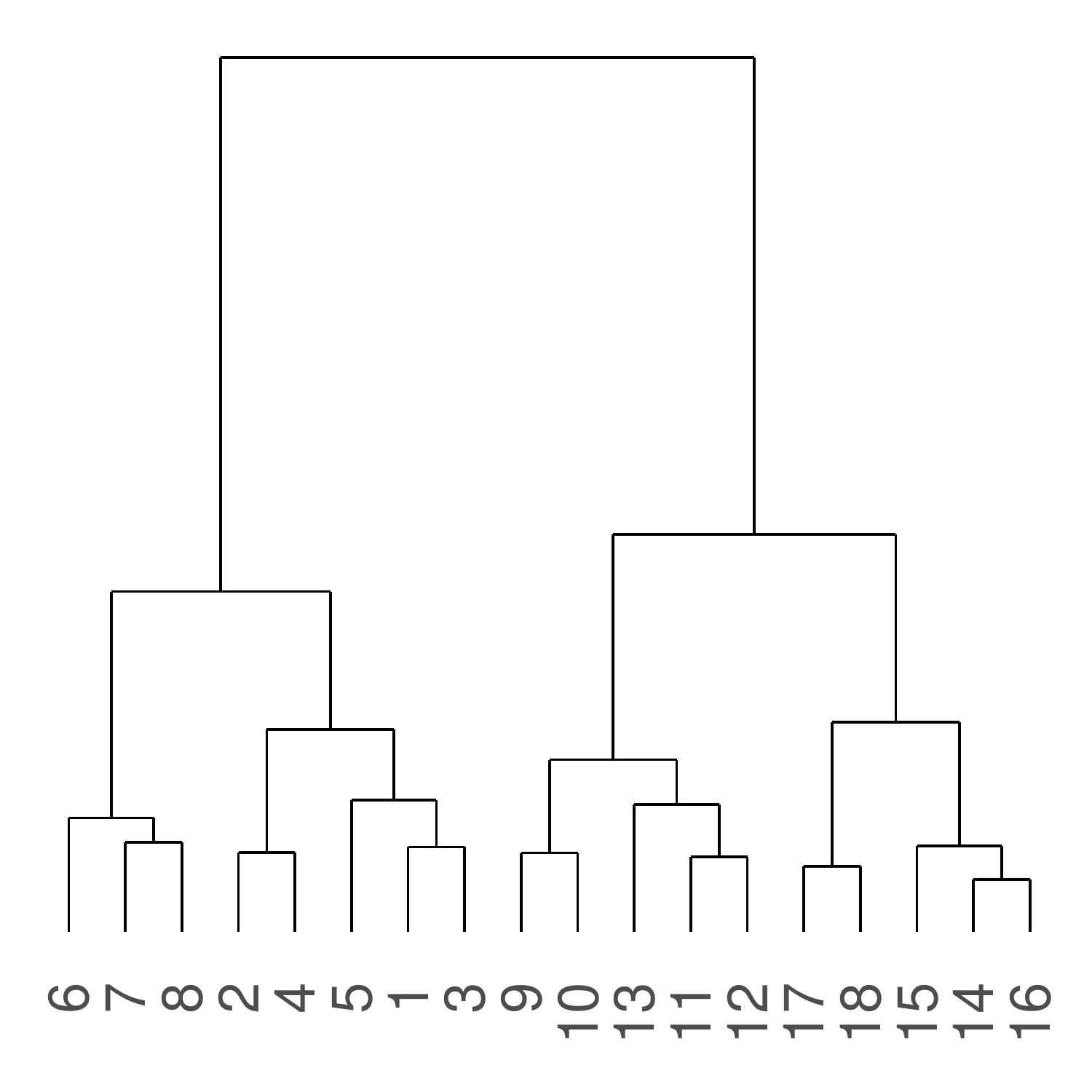} 
		\label{fig:hclust_average}} \\			
	\end{tabular}
	\caption{Hierarchical clustering of data in \Fig{scatter} under two different agglomeration methods.\label{fig:single_linkage}}
\end{figure}

There are many existing algorithms for automatically constructing partition trees, but perhaps the most often used algorithms in practice are collectively known as agglomerative hierarchical clustering methods \citep{War1963, Joh1967, LanWil1967, GowRos1969, Mur1983}.  Given a collection of points in $\Real^p$, agglomerative hierarchical clustering methods recursively merge the points which are closest together until all points are joined. 
Different choices in the definition of closeness lead to the different variants. \Fig{hclust_average} shows two trees computed by two variants of the agglomerative hierarchical clustering. 
For each tree, the eighteen points reside in the ``leaves" which are organized into a hierarchy of nested clusters that captures an increasingly coarser grouping structure as one progresses from the leaves to the root of the tree. The branch lengths in the tree quantify the similarity between pairs of points, or clusters at higher levels. We see that both trees recover binary partition trees that are similar to the ideal partition tree shown in \Fig{partition_tree}.

\subsection{Convex Hierarchical Clustering?}

Although agglomerative hierarchical methods are widely used in practice, the greedy manner in which trees are constructed often results in an unstable mapping between input data and output tree. Indeed, agglomerative hierarchical clustering methods have been shown to be highly sensitive to perturbations in the input data, namely the resulting output trees can vary drastically with the addition of a little Gaussian noise to the data \citep{Chi2017}.

One promising alternative strategy for constructing trees stably relies on formulating the clustering problem as a continuous optimization problem. Recently, several works have shown that solving a sequence of convex optimization problems can recover tree organizations \citep{PelDeSuy2005, LinOhlLju2011, HocVerBac2011, CheChiRan2015, Chi2015}. 
Given $n$ points $x_1,\ldots, x_n$ in $\Real^p$, 
we seek cluster centers (centroids) $u_i$ in $\Real^p$ attached to point $x_i$ that minimize the convex criterion
\begin{eqnarray}
\label{eq:objective_function}
E_{\gamma}(u) & = & \frac{1}{2}\sum_{i=1}^n \lVert x_i-u_i\rVert^2 + \gamma \sum_{i<j}w_{ij} \lVert u_i-u_j \rVert,
\end{eqnarray}
where $\gamma$ is a nonnegative tuning parameter, $w_{ij}$ is a nonnegative affinity that quantifies the similarity between $x_i$ and $x_j$, and $u$ is the vector in $\Real^{np}$ obtained by stacking the vectors $u_1, \ldots, u_n$ on top of each other. For now, we assume all norms are Euclidean norms; we will later consider arbitrary norms. The sum of squares data-fidelity term in \Eqn{objective_function} quantifies how well the centroids $u_i$ approximate the data $x_i$, while the sum of norms regularization term penalizes the differences between pairs of centroids $u_i$ and $u_j$. To expand on the latter, the regularization term is a composition of the group lasso \citep{YuanYi2006} and the fused lasso \citep{TibSauRos2005} and incentivizes sparsity in the pairwise differences of centroid pairs. Overall, $E_{\gamma}(u)$ can be interpreted as the energy of a configuration of centroids $u$ for a given relative weighting $\gamma$ between data-fidelity and model complexity as quantified by the regularization term. We next elaborate how $u(\gamma)$ varies as the tuning parameter $\gamma$ varies.

Because the objective function $E_{\gamma}(u)$ in \Eqn{objective_function} is strongly convex, for each value of $\gamma$ it possesses a unique minimizer $u(\gamma)$, whose $n$ subvectors in $\Real^p$ we denote by $u_i(\gamma)$. The tuning parameter $\gamma$ trades off the relative emphasis between data fit and differences between pairs of centroids. When $\gamma=0$, the minimum is attained when $u_i=x_i$, namely when each point occupies a unique cluster. As $\gamma$ increases, the regularization term encourages cluster centers to fuse together. Two points $x_i$ and $x_j$ with $u_i=u_j$ are said to belong to the same cluster. For sufficiently large $\gamma$, the $u_i$ fuse into a single cluster, namely $u_i = \overline{x}$, where $\overline{x}$ is the average of the data $x_i$ \citep{Chi2015, Tan2015}. Moreover, the unique global minimizer $u(\gamma)$ is a continuous function of the tuning parameter $\gamma$ \citep{Chi2017}; we refer to the continuous paths $u_i(\gamma)$ ,traced out from each $x_i$ to $\overline{x}$ as $\gamma$ varies, collectively as the solution path. Thus, by computing $u_i(\gamma)$ for a sequence of $\gamma$ over an appropriately sampled range of values, we hope to recover a partition tree. 

\begin{figure}[t]
    \centering
	\begin{tabular}{cc}
	    \subfloat[Gaussian Kernel Affinities]{\includegraphics[scale = 0.4]{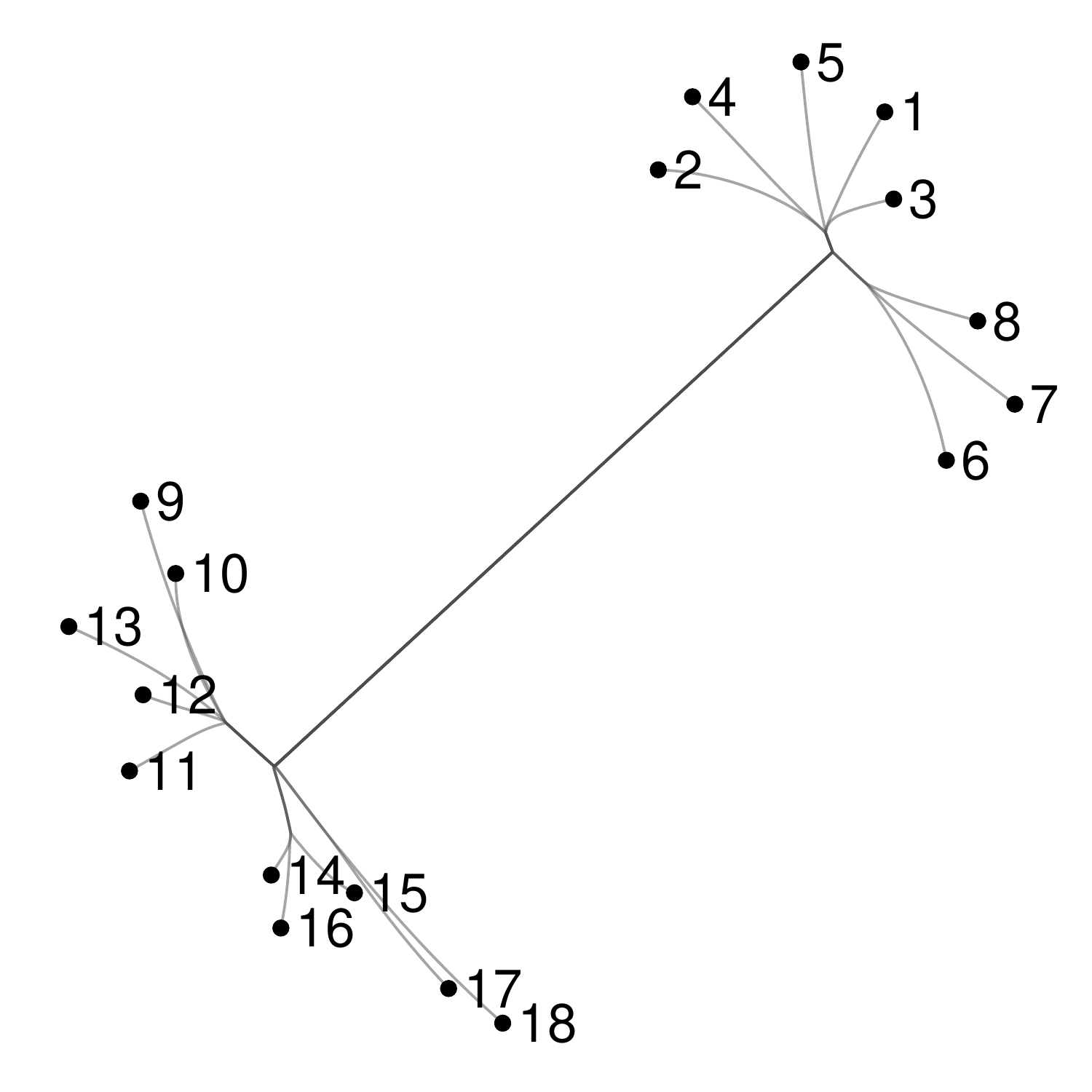} 
		\label{fig:cvxclust_gk}} 		
	    & \subfloat[Unit Affinities]{\includegraphics[scale = 0.4]{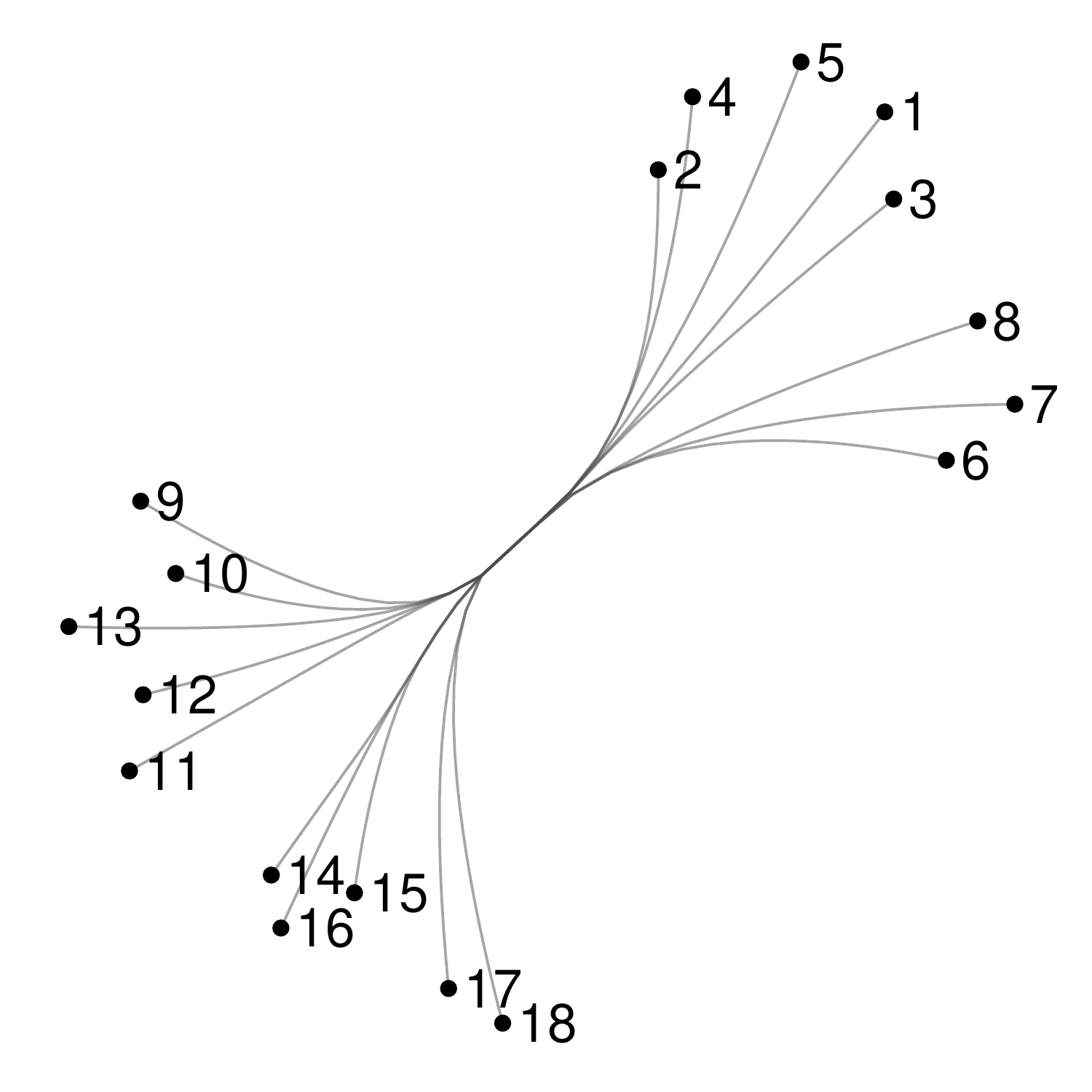} 
		\label{fig:cvxclust_unit}} \\			
	\end{tabular}
	\caption{Solution paths of convex clustering using different affinities $w_{ij}$. \label{fig:convex_cluster_paths}}
\end{figure}

\Fig{convex_cluster_paths} plots the $u_i$ as a function of $\gamma$ for two different sets of affinities $w_{ij}$.
We will discuss the differences in the recovered trees shortly, but for now we point out that computing $u(\gamma)$ for a range of $\gamma$ indeed appears to recover trees that bear similarity to the desired partition tree in \Fig{partition_tree}. Moreover, the $u_i(\gamma)$ are 1-Lipschitz functions of the data $x_i$ \citep{Chi2018}. Consequently, small perturbations to the input data $x_i$, are guaranteed to {\em not} result in disproportionately large variations in the output $u_i(\gamma)$.

At this point, the solution path of convex clustering appears to stably recover partition trees as desired. Nonetheless, questions remain as to whether convex clustering is a form of convex hierarchical clustering. Specifically, (i) when is the solution path guaranteed to produce a tree, and (ii) how do the affinities modulate the branch formation in the recovered tree?

\cite{HocVerBac2011} provide a partial answer to the first question. They prove that if unit affinities are used, namely $w_{ij} = 1$ for all $i$ and $j$, and if 1-norms are used in the regularization term in \Eqn{objective_function}, then the solution path must be a tree. On the other hand, in the same paper, they also provide an example, using the Euclidean norm in the regularization term, where the solution path can fail to be a tree. Specifically, as the tuning parameter $\gamma$ increases, it is possible for centroids to initially fuse and then ``unfuse" before eventually fusing again.

The differences in the two recovered trees shown in \Fig{convex_cluster_paths} motivate the second question. \Fig{cvxclust_gk} shows the solution path when using Gaussian kernel affinities, namely for all $i$ and $j$
\begin{eqnarray*}
w_{ij} & = & \exp\left ( -\frac{\lVert x_i - x_j \rVert_2^2}{\sigma}\right),
\end{eqnarray*}
where $\sigma$ is a positive scale parameter. Gaussian kernel affinities have been empirically shown to provide more aggressive fusion of folders closer to the leaves, and consequently more informative, hierarchical clustering results \citep{HocVerBac2011, Chi2015, Chi2017}. \Fig{cvxclust_unit} shows the solution path when using unit affinities. We see that Gaussian kernel affinities can generate a solution path that recovers the partition tree in \Fig{partition_tree}, while unit affinities can generate a solution path that recovers a less ``nested" approximation to the partition tree in \Fig{partition_tree}. The same sets of points and folders are getting shrunk together in \Fig{cvxclust_gk} and \Fig{cvxclust_unit}, but less aggressively in the latter as $\gamma$ increases.

\subsection{Contributions}

In this paper, we answer the open questions of (i) why the solution path of convex clustering can recover a tree and (ii) how affinities can be chosen to guarantee recovery of a given partition tree on the data. We first answer these questions in the case when Euclidean norms are employed in \Eqn{objective_function} and then later describe how our results can be extended to more general data-fidelity terms and arbitrary norms in the regularization term.

We clarify how the theoretical contributions in this paper differ from existing theoretical results in the convex clustering literature. \cite{Radchenko2017} present a population model for the convex clustering procedure and provide an analysis of the asymptotic properties of the sample convex clustering procedure. We note that their analysis is specific to using 1-norms in the regularization term, while we consider first the Euclidean norm before generalizing to arbitrary ones. \cite{Zhu2014} provide conditions under which two true underlying clusters can be identified by solving the convex clustering problem with appropriately chosen affinities. Similarly, \cite{She2010} and \cite{Sharpnack2012} present results when the convex clustering solution can consistently recover groupings. \cite{Tan2015} and \cite{WanZhaSun2018} present finite sample prediction error bounds for recovery of a latent set of clusters.

Our contributions differ from these prior works in two ways. First, we provide conditions on the affinities that ensure that the solution path reconstructs an {\em entire} hierarchical partition tree and clarify how these affinities can be explicitly tuned to recover a specific target tree.  With the exception of the work by \cite{Radchenko2017}, all of the other works present theoretical guarantees for recovering a {\em single} partition level rather than a nested hierarchy of partitions. 
Second, in contrast to all of the previous work, we do not make any distributional assumptions on the data. Instead, we focus in this paper on understanding the behavior of the solution path as a function of the affinities used in the regularization term. By understanding this dependency, we gain insight into why a commonly used data-driven affinities choice, namely the Gaussian kernel, works so well in practice.

\subsection{Outline}

The rest of this paper proceeds as follows. In \Sec{prelim}, we define structures needed to construct affinities that will enable us to recover a desired partition tree and once equipped with the necessary building blocks, give an overview of our main result. In \Sec{geometric_lemma}, we introduce a geometric lemma that is key to proving our main result. In \Sec{proofs}, we give proofs of the geometric lemma and our main theorem. 
In \Sec{extension}, we show how our main result can be generalized to other data-fidelity terms and regularization term norms.
In \Sec{discussion}, we conclude with a discussion on our results within the broader context of penalized regression methods for clustering.

\section{Setup and Overview of Main Result}
\label{sec:prelim}
Our main result shows that if the affinities $w_{ij}$ arise from an underlying partition tree, then that tree can be reconstructed from the solution path of the convex clustering problem. To proceed, we will need a formal definition of a partition tree and then a judicious assignment of weights to the edges in the tree graph corresponding to the partition tree.

\subsection{Partition Tree}
Let $\Omega = \left\{x_1, \dots, x_n\right\} \subset \Real^p$ be an arbitrary collection of points and let $[n]$ denote the set of indices $\{1, \ldots, n\}$. Following the notation and language employed in \citet{Ankenman2014} and \citet{Mishne2016, Mishne2017}, we say that $\mathcal{T}$ is a partition tree on the collection of points $\Omega$ consisting of $\mathcal{P}_0, \ldots, \mathcal{P}_{L}$ partitions of $\Omega$ if it has the following properties:

\begin{itemize}
\item[1.] The partition $\mathcal{P}_l = \{F_{l,1}, \ldots, F_{l,n_l}\}$ at level $l$ consists of $n_l$ disjoint non-empty subsets of indices in $\{1, \ldots, n\}$, termed folders and denoted by $F_{l,i}, i \in [n_l]$.
\item[2.] The finest partition $\mathcal{P}_0$ contains $n_0 = n$ singleton ``leaf" folders, namely $F_{0,i} = \{i\}$.
\item[3.] The coarsest partition $\mathcal{P}_{L}$ contains a single ``root" folder, namely $F_{L,1} = [n]$.
\item[4.] Partitions are nested; if $F \in \mathcal{P}_l$, then $F \subset F'$ for some $F' \in \mathcal{P}_{l+1}$, namely each folder at level $l-1$ is a subset of a folder from level $l$.
\end{itemize}
A partition tree $\mathcal{T}$ on $\Omega$ can be seen as the collection of all folders at all levels, namely $\mathcal{T} = \{F_{l,i} : 0 \leq l \leq L, i \in [n_l]\}$.

\subsection{Weighted Tree Graph}
\label{sec:weighted_tree_graph}

We next assign every folder $F_{l,i} \in \mathcal{T}$ to a node and draw an edge between nested folders in adjacent levels. Thus, if $F \in \mathcal{P}_{l}, F' \in \mathcal{P}_{l+1}$, and $F \subset F'$, then we draw an edge $(F,F')$ between $F$ and $F'$. If we let $\mathcal{E}$ denote the set of all edges between nested folders in adjacent levels, then the resulting graph $\mathcal{G} = (\mathcal{E}, \mathcal{T})$ is a tree.

\begin{figure}[th]
    \centering
	\begin{tabular}{cccc}
		\subfloat[Weighted Tree Graph]{\includegraphics[scale = 0.265]{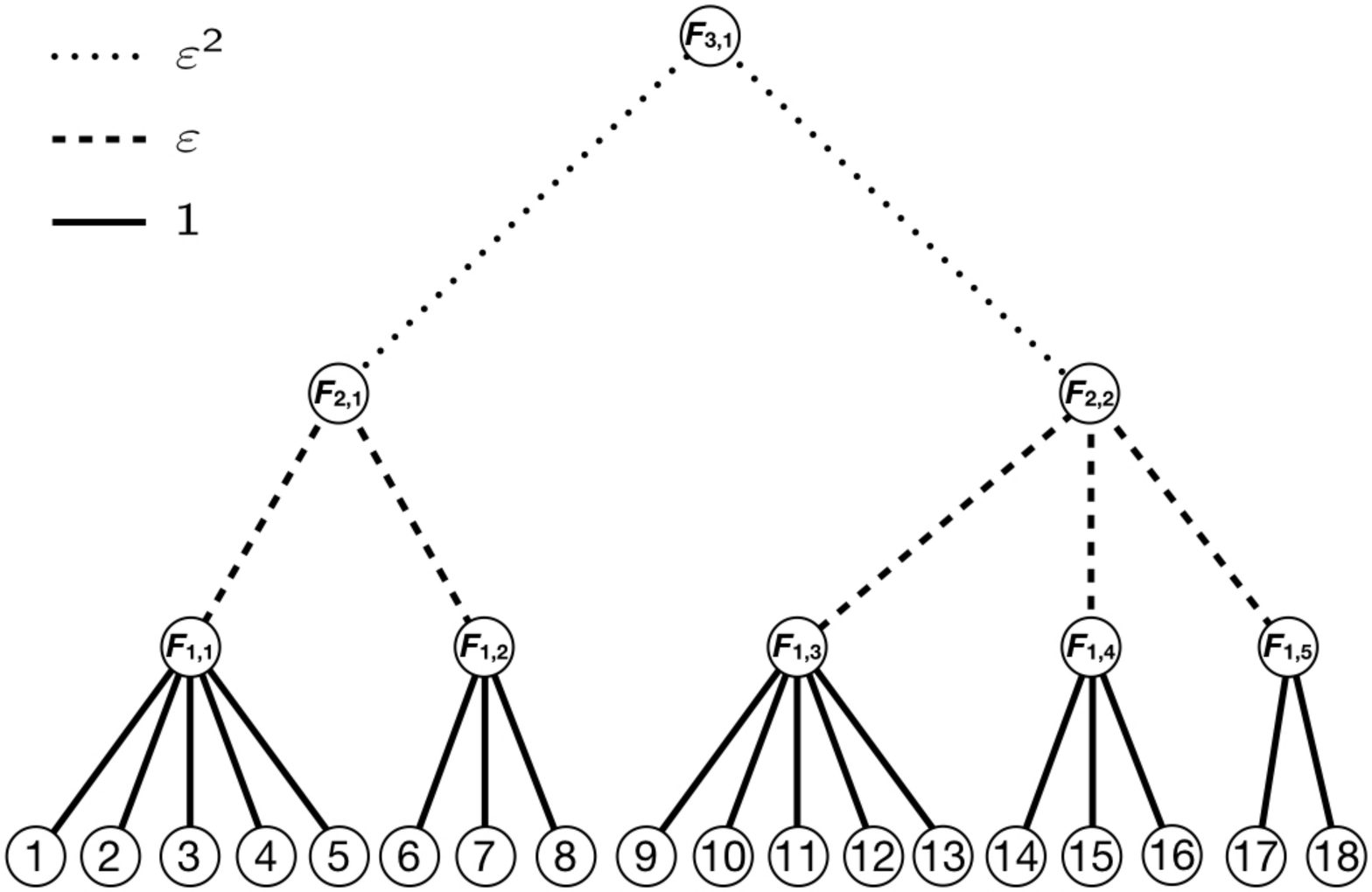}
		\label{fig:tree_null}} 
	    & \subfloat[The path $p_{15}$ from 1 to 5 produces $w_{15} = 1$.]{\includegraphics[scale = 0.265]{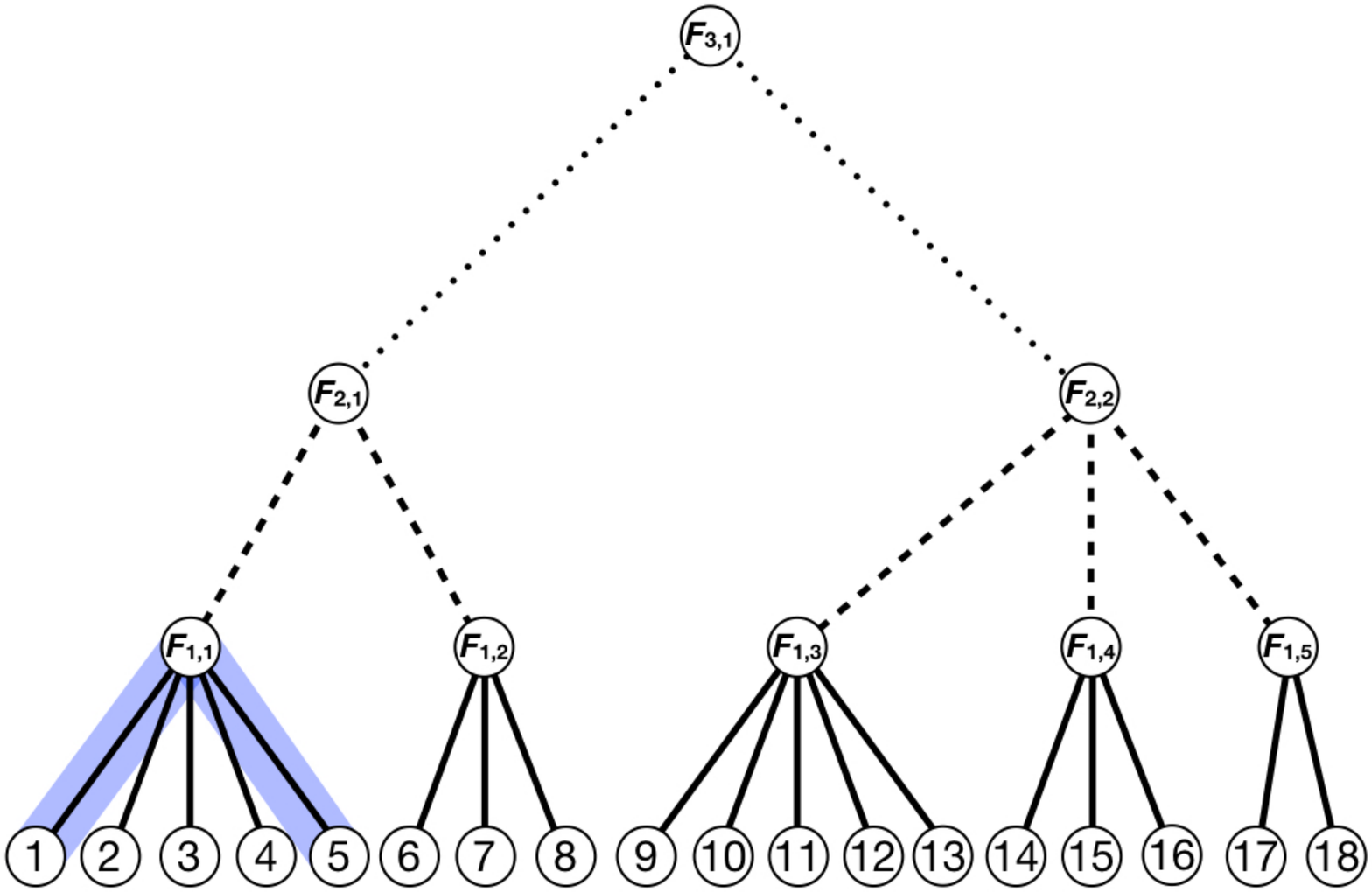} 
		\label{fig:tree_level1}} \\		
	    \subfloat[The path $p_{17}$ from 1 to 7 produces $w_{17} = \varepsilon$.]{\includegraphics[scale = 0.265]{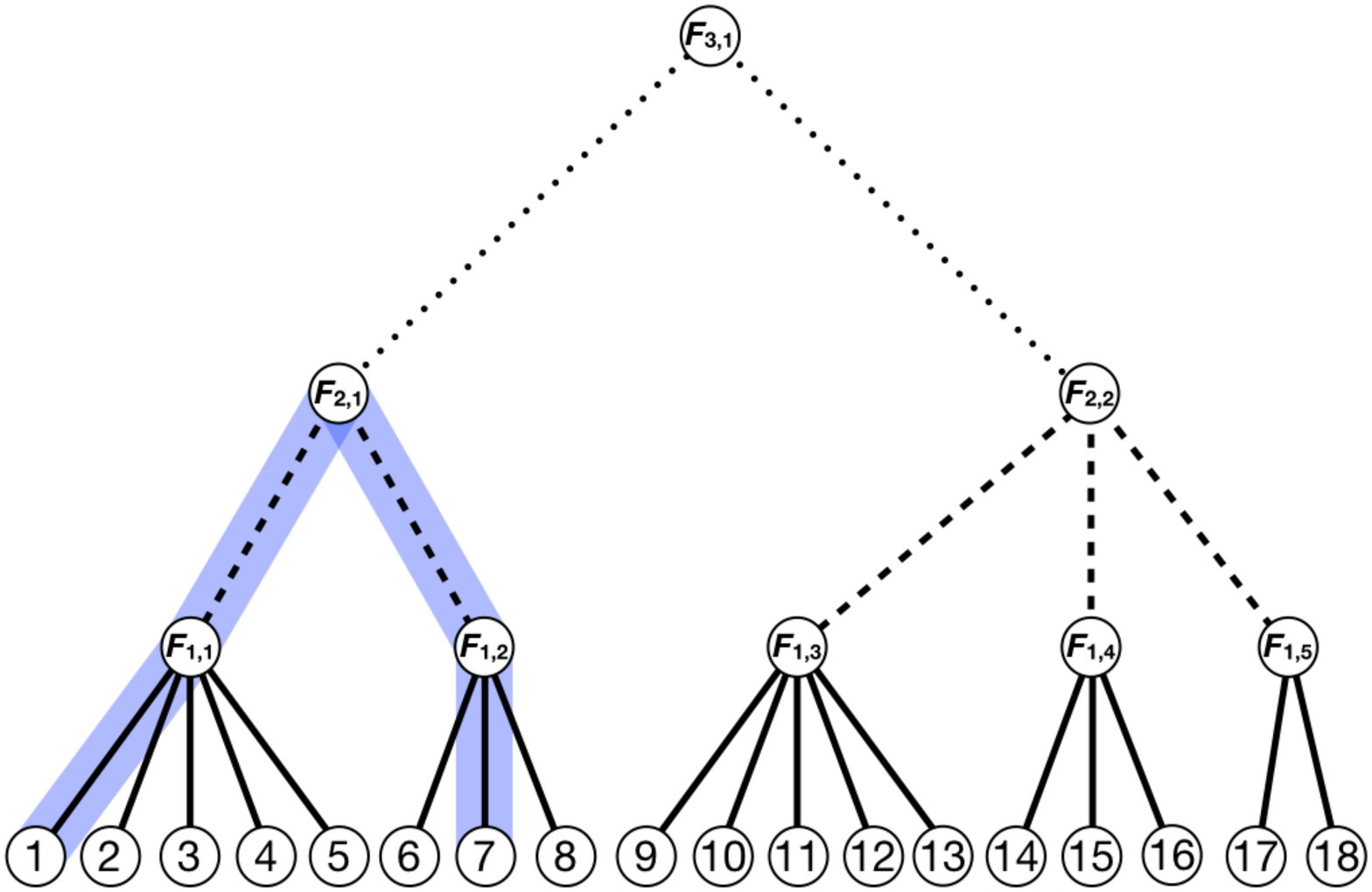} 
		\label{fig:tree_level2}} 
	    & \subfloat[The path $p_{19}$ from 1 to 9 produces $w_{19} = \varepsilon^2$.]{\includegraphics[scale = 0.265]{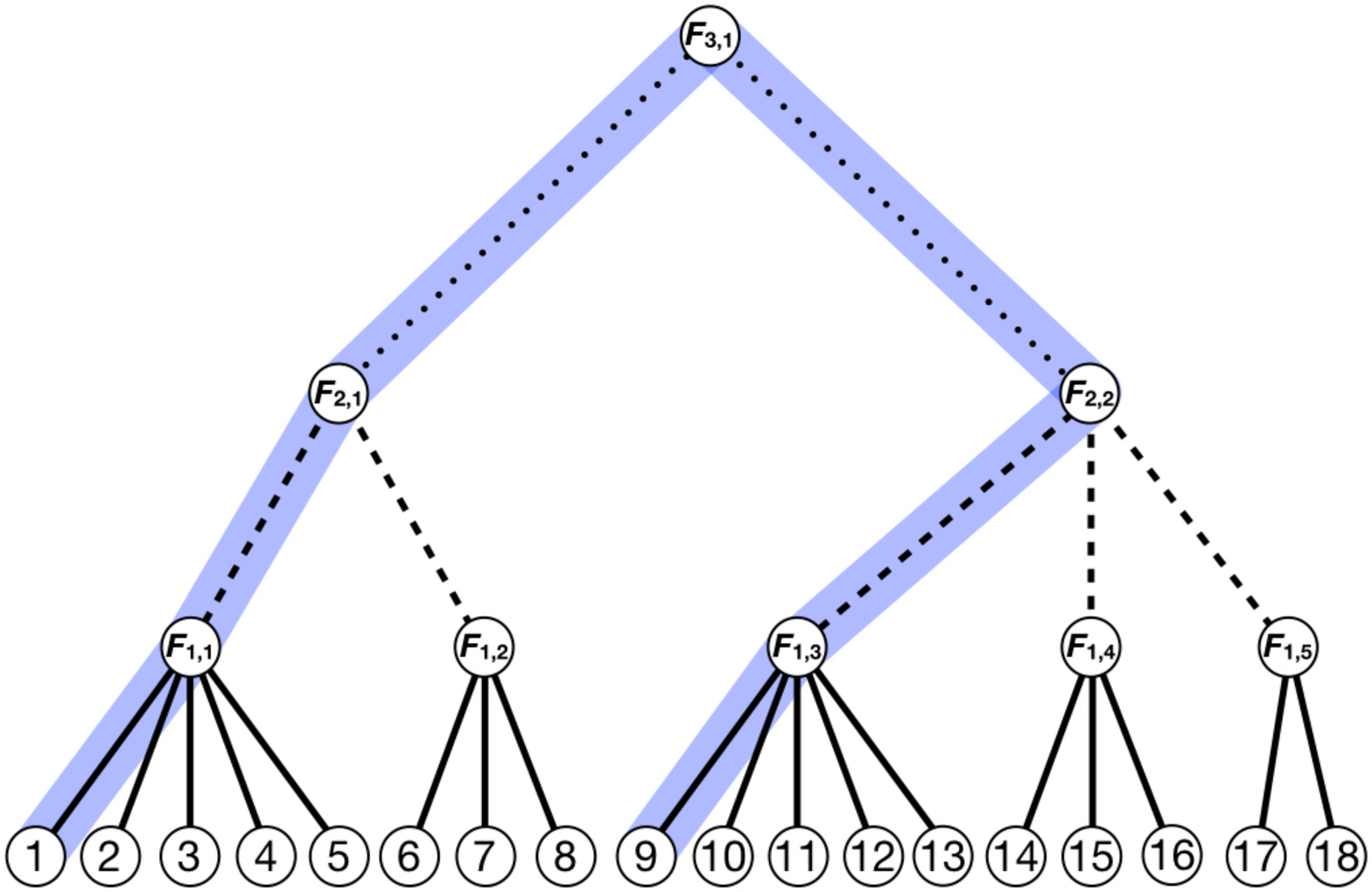} 
		\label{fig:tree_level3}} \\					
	\end{tabular}
	\caption{Weighted Tree: Edges that are solid lines have weight $1$. Edges that are dashed lines have weight $\varepsilon$. Edges that are dotted lines have weight $\varepsilon^2$. \label{fig:convex_cluster_tree}}
\end{figure}

We next assign weights on the edges in $\mathcal{E}$ as follows. Let $\varepsilon > 0$ be a fixed parameter, whose  value we will elaborate on shortly. Edges between level 0 folders and level 1 folders receive a weight of 1. Edges between level 1 folders and level 2 folders receive a weight of $\varepsilon$. Edges between level 2 folders and level 3 folders receive a weight of $\varepsilon^2$ and so on. Thus, edges between level $l$ folders and level $l+1$ folders receive a weight of $\varepsilon^l$. \Fig{tree_null} shows the weighted tree graph $\mathcal{G}$ derived from the partition tree given in \Fig{partition_tree}.

We are finally ready to construct $w_{ij}$ from the weighted tree graph. Let $F_{0,i}$ and $F_{0,j}$ be leaf nodes in the graph $\mathcal{G}$ and let $p_{ij}$ be the sequence of edges in $\mathcal{E}$ that form the path between 
$F_{0,i}$ and $F_{0,j}$. Then we set $w_{ij}$ to be the smallest weight of edges contained in $p_{ij}$. In other words, $w_{ij}$ is the smallest edge weight one sees in traveling from $i$ to $j$. \Fig{tree_level1} shows that the path $p_{15}$ from 1 to 5 in the weighted graph $\mathcal{G}$ leads to the affinity assignment $w_{15} = 1$. \Fig{tree_level2} and \Fig{tree_level3} show additional examples of how affinities are derived from the edge weights in $\mathcal{G}$. 

\subsection{Main Result}
We now state our main result.

\begin{theorem}
\label{thm:main}
There exists $\varepsilon_0 > 0$, depending on the data and the tree structure (which we assume defines the $w_{ij}$ as outlined above in \Sec{weighted_tree_graph}), so that for all $\varepsilon \in (0, \varepsilon_0)$ the
solution path
\begin{eqnarray*}
u(\gamma) & = & \underset{u_1, \dots, u_n}{\arg\min}\; \sum_{i=1}^{n}{\lVert x_i - u_i \rVert^2} + \gamma \sum_{i,j=1}^{n}{w_{ij} \lVert u_i - u_j\rVert},
\end{eqnarray*}
as parametrized by $\gamma \in (0, \gamma_0)$ traces out exactly the partition tree structure underlying the affinities $w_{ij}$ before collapsing into a point for some large, but finite, $\gamma_0$.
\end{theorem}

Informally speaking, this means that as $\gamma$ increases, elements from the same folder collapse into a single point, these folders (now single points) move themselves (or rather, the
fused points move in a coordinated manner) and then
collapse again in a way predicted by the tree (i.e.\@ folders sharing a parent folder collapse). This evolution continues on until all points have collapsed into a single point (which
happens for a finite value $\gamma_0$). We have no precise bound on the
times $\gamma$ at which these collapses happen but by making $\varepsilon_0$ sufficiently small, there is an arbitrary long time between stages of collapsing. The proof of \Thm{main} also gives a bound
on $\gamma_0$ as a byproduct.\\

\textbf{Remarks} Several additional remarks are in order.
\begin{enumerate}
\item The affinities do not need to have exactly the structure described in \Sec{weighted_tree_graph}. A more precise statement would be that there exists an $\varepsilon_0$ such that whenever we associate weight $\varepsilon_1 \in (0,\varepsilon_0)$ to the first level, then there exists an $\varepsilon$ (depending on everything and $\varepsilon_0, \varepsilon_1$) such that if we associate weight $\varepsilon_2 \in (0, \varepsilon)$ to the second level there exists an $\varepsilon_3$ (depending on everything and $\varepsilon_0, \varepsilon_1, \varepsilon_2$ etc.\@). Simply put, it suffices to have a sufficiently clear separation of scales encoded in the affinities.

Indeed, \Fig{w1k} shows the Gaussian kernel affinities $w_{1j}$ between $x_1$ and the remaining $x_j$ for $j = 2, \ldots, 18$ from the example in \Fig{scatter}. We observe clear separation of scales encoded in the Gaussian kernel affinities that align with the partition tree and corresponding weighted graph $\mathcal{G}$ in \Fig{tree_null}. Similar plots of the set of affinities associated with each data point reveal alignment with the partition tree and corresponding weighted graph $\mathcal{G}$. The key quality of the Gaussian kernel should be readily apparent, namely the Gaussian kernel naturally encodes the geometric decay in weights needed to reconstruct a partition tree embedded in Euclidean space.

\begin{figure}[t]
\centering
\includegraphics[scale = 0.4]{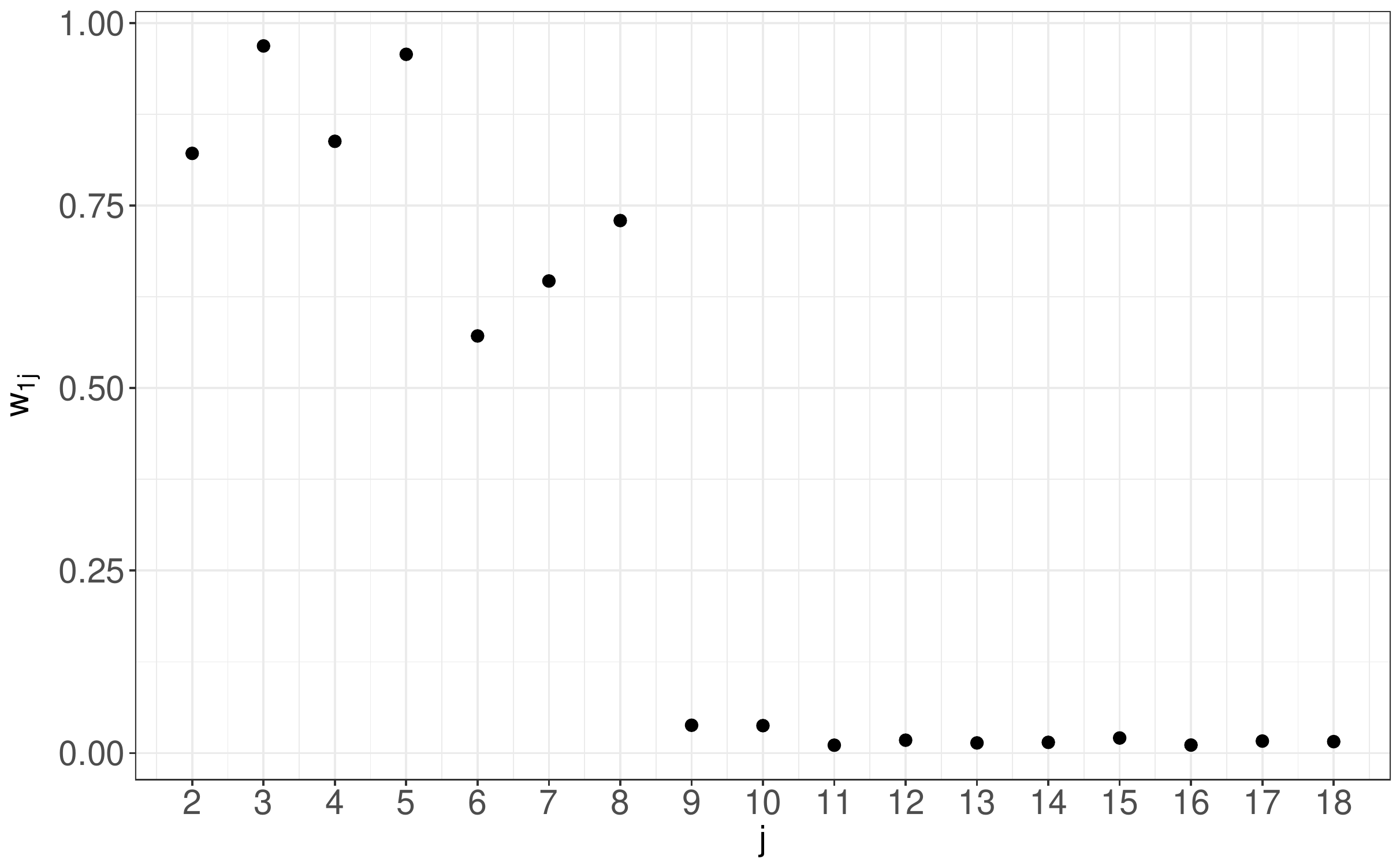}
	\caption{Gaussian kernel affinities $w_{1j}$ between $x_1$ and the other $x_j$ from the example in \Fig{scatter}.\label{fig:w1k}}
\end{figure}

\item The result is completely independent of where the $\left\{x_1, \dots, x_n\right\} \in \Real^p$ are located in space. Their location, however, affects the critical scale $\varepsilon_0$.
\item The statement guarantees that points $u_i$ fuse together with respect to the folder structure before moving to fuse with other points and their respective folder structure, however, we do not have clear control over whether they intersect in between or not. Generically, this will not happen but, for a non-generic set of $x_i$, it is possible to arrange for the $u_i$ to intersect before they fuse.
This is a consequence of our lack of conditions on the position of the points $x_i$. If the $x_i$ are located in space in a way that actually reflects the tree structure, then they will fuse upon intersecting for the first time.
\end{enumerate}

\section{A Geometric Lemma}
\label{sec:geometric_lemma}
We establish a geometric Lemma that is of intrinsic interest: it states that for any set of distinct points $\left\{u_1, \dots, u_n\right\} \in \Real^p$, one of these points $u$ (indeed, one on the boundary of the convex hull of all the points) has the property that for a suitable ``viewing direction" $v \in \Real^p$  most points are clearly visible when standing in the point $u$ and looking towards the viewing direction (in the sense of having a large inner product). We now phrase this more precisely below. Recall that the convex hull of a set $S$, denoted by $\conv S$ is the smallest convex set containing the set $S$. 

\begin{lemma}
\label{lem:geom}
For every set $S = \left\{u_1, \dots, u_n \right\} \subset \Real^p$ of $n \geq 3$ distinct points, 
there exists 
$$ u \in S \cap \partial \conv S \quad \mbox{and} \quad v \in \Real^p\quad\mbox{satisfying}~\lVert v\rVert = 1$$
such that
\begin{eqnarray}
\label{eq:property}
\frac{1}{n}\sum_{i=1 \atop u_i \neq u}^{n}{ \left\langle \frac{u_i - u}{\lVert u_i - u\rVert}, v \right\rangle} & \geq & \frac{1}{2}.
\end{eqnarray}
\end{lemma}

The statement can be summarized as follows: for a suitable point $u \in S \cap \partial \conv S$, if we map the direction to all other points onto the unit sphere $\mathbb{S}^p$, then convexity implies that there is a great circle on $\mathbb{S}^p$ such that all these directions are on one side of the great circle or on it. This can be interpreted as the dualization of the fact that there is supporting hyperplane touching the boundary of the convex hull in such a way that all of $ \conv S$ is on one side. The statement claims the existence of a boundary point $u$ such that the average projection point is bounded away from that great circle by a universal constant.

\begin{figure}[h!]
\centering
\begin{tikzpicture}[scale=1.2]
\filldraw (0,0) circle (0.05cm);
\filldraw (1,0) circle (0.05cm);
\filldraw (0,1) circle (0.05cm);
\filldraw (1.2,1.2) circle (0.05cm);
\filldraw (2,-1) circle (0.05cm);
\filldraw (2,2) circle (0.05cm);
\filldraw (-1,0) circle (0.05cm);
\filldraw (-0.2,0.4) circle (0.05cm);
\filldraw (0.4,-0.5) circle (0.05cm);
\filldraw (-0.2,-0.2) circle (0.05cm);
\filldraw (-0.3,0.5) circle (0.05cm);
\filldraw (0.4,0.3) circle (0.05cm);
\filldraw (0,0) circle (0.05cm);
\draw [thick] (-1,0) -- (0,1) -- (2,2) -- (2,-1) -- (0.4,-0.5) -- (-1,0);
\draw [->, thick] (2,-1) -- (1.5, -0.5);
\node at (2.3, -1) {$u$};
\node at (1.6, -0.4) {$v$};
\end{tikzpicture}
\caption{A set of points in $\Real^2$: there exists a point $u$ on the boundary of the convex hull and a direction $v$ such that the average inner product of $(u_i - u)/\lVert u_i - u\rVert$ and $v$ is bounded away from 0 by a universal constant.}
\end{figure}
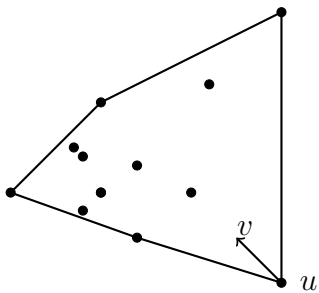

We will use \Lem{geom} to study the regularization term in \Eqn{objective_function}, namely the functional
\begin{eqnarray*}
J(u) & = & \sum_{i,j =1}^{m}{ \lVert u_i - u_j \rVert } \qquad \mbox{for a given set of distinct points}~\left\{u_1, u_2, \dots, u_m\right\} \subset \Real^p.
\end{eqnarray*}
The functional $J$ is clearly minimized for any collection of $u_i$ that are all identical. Consequently, any collection of distinct $u_i$ represents a suboptimal configuration of centroids and therefore admits a descent direction that leads to a decrease in energy. The power of \Lem{geom} is that it identifies a direction that guarantees a large amount of decrease in $J$. To see this, we write down the directional derivative of $J$ explicitly.

The directional derivative of moving $u_j$ in direction $v \in \Real^p$, normalized to $\lVert v\rVert = 1$ is computed as
\begin{equation}
\label{eq:directional_derivative}
\begin{split}
 \left\langle \frac{\partial J}{\partial u_j}, v \right\rangle & \amp = \amp \lim_{t \rightarrow 0} \frac{1}{t} \sum_{i \neq j}{\lVert u_i - (u_j + t v)\rVert - \lVert u_i - u_j\rVert} \\
& \amp = \amp  \lim_{t \rightarrow 0} \frac{1}{t} \sum_{i \neq j}{  \sqrt{  \big\langle u_i - (u_j + t v), u_i - (u_j + t v) \big\rangle} - \lVert u_i - u_j \rVert}  \\
& \amp = \amp  \sum_{i \neq j}{\lim_{t \rightarrow 0} \frac{1}{t} \left( \sqrt{\lVert u_i - u_j \rVert^2 - 2t  \left\langle u_i - u_j , v \right\rangle + t^2}  - \lVert u_j - u_i \rVert \right)}\\
& \amp = \amp  -\sum_{i \neq j}{\left\langle  \frac{ u_i - u_j  }{ \lVert u_i - u_j \rVert }, v \right\rangle }.
\end{split}
\end{equation}
The expression for the directional derivative given in (\ref{eq:directional_derivative}), in conjunction with \Lem{geom}, shows that it is always possible to find one point such that moving it $\varepsilon$ in a certain direction decreases the entire functional by at least $(n/3)\varepsilon$. This may be understood as a non-degeneracy condition on the functional $J$. The existence of a direction of guaranteed minimum decrease in $J$ will be essential in proving \Thm{main}. \\

Before proceeding to proofs of \Lem{geom} and \Thm{main}, we also note the following consequence, whose proof is in the Appendix.
\begin{corollary}
\label{cor:more_geometry} Let $S = \left\{u_1, \dots, u_n \right\} \subset \Real^p$ be a set of distinct points. Then there exist at least $n/6$ points $u \in S$ having the property that for some $\lVert v \rVert= 1$
\begin{eqnarray*}
\frac{1}{n}\sum_{i=1 \atop u_i \neq u}^{n}{ \left\langle \frac{u_i - u}{\lVert u_i - u \rVert}, v \right\rangle} & \geq & \frac{1}{4}.
\end{eqnarray*}
\end{corollary}

This simple statement has non-trivial implications: the geometric Lemma may seem like these vantage points from which to observe the entirety of the set without having too many small inner products are rare.  To the contrary, \Cor{more_geometry} declares that the property is surprisingly common and enjoyed by a universal fraction of all points. We believe this result to be of substantial independent interest since it can be interpreted as a basic statement (with universal constants) in a general Hilbert space. It could be of interest to further pursue this line of investigation.

\section{Proofs}
\label{sec:proofs}
We now prove \Lem{geom} and \Thm{main}.

\subsection{Geometric Lemma}

\begin{proof} 
Let $S = \left\{u_1, u_2, \dots, u_n\right\}$. Select an arbitrary $u \in \partial S \cap \conv S$, and let $y \in S$ be a point in the set furthest from $u$ (there may be more than one such point), formally
\begin{eqnarray}
\label{eq:distance}
\| u - y \| & = &  \max_{1 \leq i \leq n}{\| u - u_i\|}
\end{eqnarray}
It is easy to see that $y$ resides on the boundary of the convex hull; $y$ is in fact an extreme point. We now show that $u$, equipped with
the viewing direction vector $v_1 = (y-u)/\lVert y-u \rVert$, or $y$, equipped with the viewing direction vector $v_2 = -v$, has the desired property. We first show that for every  $u_i \notin \left\{u, y\right\}$

\begin{eqnarray}
\label{eq:inequality}
\left\langle \frac{u_i - u}{\lVert u_i - u\rVert}, v_1 \right\rangle +  \left\langle \frac{u_i - y}{\lVert u_i - y\rVert}, v_2 \right\rangle & \geq & 1.
\end{eqnarray}
Since we are only dealing with three points $u, y,$ and $u_i$, all angles are determined by the corresponding triangle, which we can assume without loss of generality to reside in $\Real^2$. Moreover, the invariance under dilation, translation and rotation enables us to assume that $u = (0,0)$ and $y=(1,0)$. If
we write $u_i = (a,b)$, then the expression on the left hand side of (\ref{eq:inequality}) simplifies to
\begin{eqnarray}
\label{eq:equality}
\left\langle \frac{u_i - u}{\lVert u_i - u \rVert}, v_1 \right\rangle +  \left\langle \frac{u_i - y}{\lVert u_i - y \rVert}, v_2 \right\rangle & = & \frac{a}{\sqrt{a^2+b^2}} + \frac{1-a}{\sqrt{(1-a)^2 + b^2}},
\end{eqnarray}
and the condition on the distances $\lVert u - u_i \rVert$ and $\lVert y - u_i \rVert$ required by (\ref{eq:distance}) implies that
\begin{eqnarray}
\label{eq:constraint}
\max\left\{ a^2 + b^2, (1-a)^2 + b^2\right\} & \leq & 1.
\end{eqnarray}
Minimizing the expression in (\ref{eq:equality}) subject to the constraint in (\ref{eq:constraint}) gives us the desired inequality in (\ref{eq:inequality}); almost equality is attained for $u_i$ very close to either $u$ or $y$ (and as orthogonal as possible to $v$) and equality is attained for $(a,b)=(1/2, \sqrt{3}/2)$. We then sum the left and right hand sides of (\ref{eq:inequality}) over $i = 1, \ldots, n$ to arrive at the inequality
\begin{eqnarray}
\label{eq:inequality2}
\sum_{i=1 \atop u_i \neq u}^{n}{\left\langle \frac{u_i - u}{\lVert u_i - u \rVert}, v_1 \right\rangle} + \sum_{i=1 \atop u_i \neq y}^{n}{ \left\langle \frac{u_i - y}{\lVert u_i - y \rVert}, v_2 \right\rangle } & \geq & n,
\end{eqnarray}
which follows from realizing that each of the sums contains one term that is equal to 1 and that the remaining sum runs over all $u_i \notin \left\{u, y\right\}$ yielding at least a total of $n-2$.
Thus at least one of the two terms is size $n/2$ and we obtain the desired result.
\end{proof}

\subsection{Main Theorem}
\textbf{Outline}
The proof is based on the self-similarity of the statement. We essentially show that points at the lowest level fuse in the right way with points in the same leaves (those
who have mutual affinity 1). Once they are fused, we show that they stay fused for all subsequent values of $\gamma$. The newly emerging problem turns out to be exactly of the
same type as the original one: we re-interpret fused points as single points with a mutual interaction now at scale $\sim \varepsilon$ (which becomes the dominant scale since points
with $w_{ij} = 1$ are already fused). At every step, the arguments will go through provided $\varepsilon$ is sufficiently small (but positive) and since the tree is of finite height, the result
follows. To be more precise, the argument will proceed as follows.

\begin{enumerate}
\item We assume that the $x_i$ are fixed and that the $u_i$ are solutions of the minimization problem
$$\inf_{u_1, \dots, u_n} \left[ \sum_{i=1}^{n}{\lVert x_i - u_i \rVert^2} + \gamma \sum_{i,j=1}^{n}{w_{ij} \lVert u_i - u_j \rVert} \right].$$
Plugging in an example shows that the minimal energy is uniformly bounded in $\gamma$. This has some basic
implications: the $u_i$ cannot be too far away from the $x_i$ and not too far away from each other.
\item We then study a subset of points $\left\{x_1, \dots, x_n\right\}$ contained in a leaf of the tree. This means that
their mutual affinity satisfies $w_{ij} = 1$ and the affinity between any of these points to any other point not in the leaf of the partition is at most $\varepsilon$.
\item We then focus exclusively on these point sets and prove that for $\gamma$ sufficiently large, these sets are necessarily fused in a point. This is where \Lem{geom} will be applied.
\item This contradiction proves that for $\gamma$ sufficiently large, the point sets in the leaf are fused into exactly one point as desired. Once this has been shown,
the full statement essentially follows by induction since these fused points interact exactly as individual points used to do; having common parents in the tree
becomes the next-level analogue of being associated to the same leaf. The result then follows.
\end{enumerate}

\begin{proof}
We introduce the energy of the minimal energy configuration for $\gamma > 0$ as
\begin{eqnarray*}
E(\gamma) & = & \underset{u}{\inf}\; E_\gamma(u) =\underset{u}{\inf} \left[ \sum_{i=1}^{n}{\lVert x_i - u_i \rVert^2} + \gamma \sum_{i < j}{w_{ij} \lVert u_i - u_j \rVert} \right].
\end{eqnarray*}
By setting $u_1 = u_2 = \dots = u_n$ and putting these points in the center of mass of $\left\{x_1, \dots, x_n\right\}$, we observe that this energy is uniformly bounded for all $\gamma$
\begin{eqnarray*}
E & = & \sup_{\gamma>0}~ E(\gamma) \amp \leq \amp \sum_{i=1}^{n}{ \left \lVert x_i - \frac{1}{n} \sum_{i=1}^{n}{x_i} \right\rVert^2} \amp < \amp \infty.
\end{eqnarray*}
We decompose the energy functional $E(\gamma)$ as
\begin{eqnarray}
\label{eq:decomposition}
E(\gamma) & = & E_1(\gamma) + E_2(\gamma),
\end{eqnarray}
where
\begin{eqnarray*}
E_1(\gamma) & = & \sum_{i=1}^{n}{\lVert x_i - u_i \rVert^2} + \gamma \sum_{(i,j) \in \mathcal{E}_1} \lVert u_i - u_j\rVert,
\end{eqnarray*}
where $\mathcal{E}_1 = \{ (i,j) : w_{ij} = 1\}$ and
\begin{eqnarray*}
E_2(\gamma) & = & \gamma \sum_{(i,j) \in \mathcal{E}_2} w_{ij} \lVert u_i - u_j \rVert,
\end{eqnarray*}
where $\mathcal{E}_2 = \{ (i,j) : w_{ij} \leq \varepsilon < 1\}$. The decomposition (\ref{eq:decomposition}) makes explicit that, for $\varepsilon$ sufficiently small, the functional $E_2(\gamma)$ can be interpreted as an error term, while the dominant dynamics are determined by $E_1(\gamma)$. 
We now claim that for $\gamma$ sufficiently large (where sufficiently large depends on everything except the parameter $\varepsilon$) any subset of the points $u_i$ whose mutual affinities are 1 (i.e.\@ all the members of one of the leaves in the tree) are fused in a point. The argument can be made quantitative and
\begin{eqnarray*}
\gamma & > & \frac{8\sqrt{E}}{n}
\end{eqnarray*}
will turn out to be sufficient. 

We will invoke \Lem{geom} shortly, but in order to do so we need to ensure that all points are distinct. It is easy to see that the energy $E$ is a continuous functional, this means that we can move any potentially clumped points apart by accepting an arbitrarily small increase of energy; the remainder of the argument works as follows: if points happen to be clumped together -- but not in exactly one point but in several -- then we may move all of them an arbitrarily small bit. We can accept an arbitrarily small increase of energy as long as we are able to then deduce a definite decrease in energy afterwards (that will depend on the diameter of the $u_i$); this contradiction shows that the clumping has to occur in exactly one point.
The next step in the argument is dynamical: we compute the effect of moving one of the points an infinitesimal amount (this is already using the assumption that all $u_i$ are distinct). 
Reusing the computation in \Eqn{directional_derivative}, we see that
\begin{eqnarray}
\label{eq:perturb}
\left \langle \frac{\partial E }{\partial  u_j}, v \right \rangle & = & 2\left \langle u_j - x_j , v \right \rangle - \gamma \sum_{i =1 \atop i \neq j}^{n}{\left\langle  \frac{ u_i - u_j  }{ \lVert u_i - u_j \rVert }, v \right\rangle } + \left \langle \frac{\partial  }{\partial  u_j}   \gamma \sum_{(i,j) \in \mathcal{E}_2} w_{ij} \lVert u_i - u_j \rVert, v \right \rangle.\qquad
\end{eqnarray}
The first term on the right hand side of (\ref{eq:perturb}) is bounded above by
\begin{eqnarray}
\label{eq:bound4}
 2 \left \lvert \left \langle u_j - x_j , v \right \rangle \right \rvert & \leq & 2\lVert x_j - u_j\rVert \amp \leq \amp 2 \sqrt{E},
\end{eqnarray}
and the third term on the right hand side of (\ref{eq:perturb}) is bounded above by 
\begin{eqnarray}
\label{eq:bound5}
\left\lVert \frac{\partial  }{\partial  u_j}   \gamma \sum_{(i,j) \in \mathcal{E}_2} w_{ij} \lVert u_i - u_j \rVert \right\rVert & = &
\gamma \left \lVert \sum_{i: (i,j) \in \mathcal{E}_2} w_{ij} \frac{u_i - u_j}{\lVert u_i - u_j \rVert} \right \rVert
\amp \leq \amp
\gamma \varepsilon  n.
\end{eqnarray}

\Lem{geom} guarantees that there exists $u_j$ for which the second term on the right hand side of (\ref{eq:perturb}) is smaller than $-\gamma n/2$. The proof of \Lem{geom} is even stronger and guarantees that if $\lVert u_i - u_j\rVert = \diam\left\{u_1, \dots, u_n\right\}$, then either $u_i$ or $u_j$ has the desired property and can be moved in a suitable direction $v$. Plugging the $u_j$ and $v$ from \Lem{geom} into both sides of (\ref{eq:perturb}) and applying inequalities (\ref{eq:bound4}) and (\ref{eq:bound5}), we arrive at the following inequality.
\begin{eqnarray}
\label{eq:guaranteed_decay}
\left \langle \frac{\partial E }{\partial  u_j}, v \right \rangle & \leq & D(\gamma) \amp = \amp 2 \sqrt{E} - \gamma n\left( \frac{1}{2} - \varepsilon \right).
\end{eqnarray}

The inequality in (\ref{eq:guaranteed_decay}) guarantees that for $\gamma$ sufficiently large (depending on $E, \varepsilon$, and $n$), there exists a descent direction $v$ with a decrease rate of at least $D(\gamma)$. We now move the point $u_j$ with the desired property in the descent direction $v$ or a direction very close to it if we are in danger of colliding with another point (this is allowed because of the continuity of all functionals involved and the fact that we do not work with sharp constants). We do this until the moved point does not have the desired property anymore. Then another point will have the desired property, and we can repeat the procedure. It is easily seen that the total decrease in $E$ obtained this way is at least  $D(\gamma) \cdot \diam P$, where $P$ is the set of $u_j$ belonging to a leaf folder.

This, however, implies that as soon as we can guarantee the existence of a strict decent direction, we are not dealing with a minimizer unless the diameter is 0 and all points are fused.
This argument can be made quantitative and assuming $\varepsilon \leq 1/4$, we see that $D(\gamma)$ is negative for all
\begin{eqnarray*}
\gamma & \geq & \frac{8 \sqrt{E}}{n}.
\end{eqnarray*}
This shows that all the points in the leaf have to have fused into a single point for some $\gamma$ less than $8 \sqrt{E}/n$ and will stay in this configuration for all subsequently larger $\gamma$.
A careful inspection of the proof shows that we do not require $w_{ij} =1$ for points in the same partition: it suffices if $1 \leq w_{ij} \leq c$ for some constant $c$
if subsequent parameter choices of $\gamma$ are allowed to depend on that. The full
statement now follows by induction: points in leaves become a single point, their parent structure determines the next collection of leaves and the product of their affinities determines the new affinities.
\end{proof}

\section{Extensions of the Main Theorem}
\label{sec:extension} The proof of \Thm{main} relies on rather elementary analysis and consequently is quite flexible. Indeed, the proof can be immediately extended to more general notions of energy of the type
\begin{eqnarray*}
E_{\gamma}(u) & = &  \phi(x_1, \dots, x_n, u_1, \dots, u_n) + \gamma \sum_{i < j}{w_{ij} \left\lVert  u_i - u_j \right\rVert_{X}},
\end{eqnarray*}
where $X$ is an arbitrary norm on $\Real^p$ and $\phi$ is assumed to satisfy the following properties:
\begin{enumerate}
\item The function $\phi: \Real^{p \times n} \rightarrow \Real_{\geq 0}$ enforces some degree of data-fidelity and compactness.
\item For all $u$ for which
\begin{eqnarray*}
\phi(x_1, \dots, x_n, u_1, \dots, u_n) + \gamma \sum_{i,j=1}^{n}{w_{ij} \left\lVert  u_i - u_j \right\rVert_{X}} & \leq &  \inf_{x \in \Real^p} \phi(x_1, \dots, x_n, x, \dots, x),
\end{eqnarray*}
we have
\begin{eqnarray*}
\left\lVert  \frac{\partial}{\partial u_i} \phi(x_1, \dots, x_n, u_1, \dots, u_n) \right\rVert & \leq & c
\end{eqnarray*}
where $c$ only depends on $\gamma$ and $\left\{x_1. \dots, x_n\right\}$.
\end{enumerate}

The argument proceeds in exactly the same way and makes crucial use of the fact that any two norms in a finite-dimensional Euclidean space are equivalent up to constants, namely
\begin{eqnarray*}
c_5 \lVert x\rVert_{\ell^2} & \leq & \lVert x \rVert_{\scaleto{X}{4.5pt}} \amp \leq \amp c_6 \lVert x\rVert_{\ell^2}.
\end{eqnarray*}
Since constants can always be absorbed in $\gamma$, this reduces to our case, namely $X = \ell^2$. \\

\begin{proof}(Sketch of the argument)
Setting all $u_i = x$ and minimizing over $x$ implies that the energy is uniformly bounded in $\gamma$ (with a bound depending only on $\left\{x_1, \dots, x_n \right\}$). Since the norm $X$ is comparable to the Euclidean norm, this implies that any minimizing configuration $\left\{u_1, \dots, u_n\right\}$ has to have a bounded diameter (with a bound depending only on $\left\{x_1, \dots, x_n \right\}$). Then, for $\gamma$ sufficiently large (depending on $c$), \Lem{geom} implies a direction of decay and thus points are eventually fused. We leave the precise details to the interested reader.
\end{proof}

We close this section by noting that the generality of our result opens the door to intriguing applications. For example, one potential application of our extension is to construct partition trees of regression coefficients in clustered regression \citep{Bondell2008, She2010, Witten2014, KeFanWu2015}. We leave these investigations as future work.

\section{Discussion}
\label{sec:discussion}

In this paper, we answered the question of when the convex clustering solution path recovers a tree. The key to ensuring the recovery of a well nested partition tree is the use of affinities that encourage the fusions within a folder before fusions with higher level folders and so on as the tuning parameter $\gamma$ increases. By choosing the edge weight parameter $\varepsilon$ sufficiently small, different folders have very little incentive to interact, and the optimization problem is essentially decoupled. As $\gamma$ increases, the same procedure repeats itself.

We end with a discussion on the relationship between convex and non-convex formulations of penalized regression based clustering. Although we focus in this paper on the ability of convex clustering to recover a potentially deep hierarchy of nested folders, our result also sheds light on a gap in theory and practice that convex clustering's performance can be significantly improved when using non-uniform data-driven affinities when seeking a shallow or single level of nested folders. In practice, Gaussian kernel affinities have been observed to work well, but these affinity choices have until now lacked formal justification.

Indeed, non-uniform affinities provide the link between convex clustering and other penalized regression-based clustering methods that use folded concave penalties. It is well known that 1-norm penalties lead to biased parameter estimates. Bias is the price for simultaneously performing variable selection and estimation. In the context of convex clustering, the centroid estimates $u_i$ are biased towards the grand mean $\overline{x}$. Consequently, others have proposed employing a folded concave penalty instead of a norm in the regularization terms  \citep{PanShenLiu2013,Marchetti2014,Wu2016}.  Folded concave penalties suffer far less bias in exchange for giving up convexity in the optimization problem, which means that iterative algorithms can typically at best converge only to a KKT point.

Suppose we were to employ a folded concave penalty, such as the smoothly clipped absolute deviation \citep{FanLi2001} or minimax concave penalty \citep{Zhang2010}, and seek to minimize the following alternative objective to \Eqn{objective_function}
\begin{eqnarray}
\label{eq:folded_concave_objective}
\tilde{E}_\gamma(u) & = & \frac{1}{2} \sum_{i=1}^n \lVert x_i - u_i \rVert^2 + \gamma \sum_{i < j}\varphi \left(\lVert u_i - u_j\rVert\right),
\end{eqnarray}
where each $\varphi : [0,\infty) \mapsto [0, \infty )$ has the following properties: (i) $\varphi$ is concave and differentiable on $(0, \infty)$, (ii) $\varphi$ vanishes at the origin, and (iii) the directional derivative of $\varphi$ exists and is positive at the origin.

Since $\varphi$ is concave and differentiable, for all positive $z$ and $\tilde{z}$
\begin{eqnarray*}
	\varphi(z) & \leq & \varphi(\tilde{z}) + \varphi'(\tilde{z})(z - \tilde{z}).
\end{eqnarray*}
In other words, the first order Taylor expansion of a differentiable concave function $\varphi$ provides a tight global upper bound at the expansion point $\tilde{z}$. Thus, we can construct a function that is a tight upper bound of the function $\tilde{E}_\gamma(u)$
\begin{eqnarray}
\label{eq:majorization}
g_\gamma(u \mid \tilde{u}) & = & 
\frac{1}{2} \sum_{i=1}^n \lVert x_i - u_i \rVert^2 + \gamma \sum_{i < j}w_{ij} \lVert u_i - u_j\rVert+ c_7,
\end{eqnarray}
where $c_7$ is a constant that does not depend on $u$, and $w_{ij}$ are affinities that depend on $\tilde{u}$, namely
\begin{eqnarray*}
w_{ij} & = & \varphi'\left(\lVert \tilde{u}_i - \tilde{u}_j \rVert \right).
\end{eqnarray*}
Note that if we take $\tilde{u}_i$ to be the data $x_i$, and $\varphi(z)$ to be  the following variation on the error function
\begin{eqnarray*}
\varphi(z) & = & \int_{0}^z e^{-\frac{\alpha^2}{\sigma} }d\alpha,
\end{eqnarray*}
then the bounding function given in \Eqn{majorization} coincides, up to an irrelevant shift and scaling, with the convex clustering objective using Gaussian kernel affinities. 

The function $g_\gamma(u \mid \tilde{u})$ is said to majorize the function $\tilde{E}_\gamma(u)$ at the point $\tilde{u}$  \citep{LanHunYan2000} and minimizing it corresponds to performing one step of the local linear-approximation algorithm \citep{Zou2008, Schifano2010}, which is a special case of the majorization-minimization (MM) algorithm \citep{LanHunYan2000}.  Thus, we can see that employing Gaussian kernel affinities corresponds to taking one step of a local linear-approximation algorithm applied to a penalized regression based clustering with an appropriately chosen folded concave penalty.

In practice, 
variants that employ folded concave penalties take multiple steps of the local linear approximation. So at the $k$th step,
\begin{eqnarray*}
u^{(k)} & = & \underset{u}{\arg\min}\; \frac{1}{2} \sum_{i=1}^n \lVert x_i - u_i \rVert^2 + \gamma \sum_{i < j}
\varphi'\left(\lVert u^{(k-1)}_i - u^{(k-1)}_j \rVert \right)
 \lVert u_i - u_j\rVert.
\end{eqnarray*}
As affinities represent a data-driven way to approximate the partition tree, one can see that employing folded concave penalties corresponds to implicitly recomputing the affinities, which corresponds to refining our estimate of the partition tree based on the data.

In light of this current work, this last observation raises two interesting questions: (i) what partition tree is being recovered by a solution path of a penalized regression-based clustering method that uses a folded concave penalty and (ii) when is the recovered partition tree substantially different than the tree corresponding to a one-step local linear approximation? We leave these questions to future work.

\begin{center}
{\large\bf ACKNOWLEDGMENTS}
\end{center}
We thank Raphy Coifman for pointing out \Cor{more_geometry}.

\appendix

\section{Proof of \Cor{more_geometry}}
\label{sec:more_geometry}

\begin{proof} \Lem{geom} guarantees the existence of a point $u$, call it $\tilde{u}_{1}$, and viewing direction vector $v_1$ that satisfies inequality (\ref{eq:property}). Remove $\tilde{u}_1$ from the set $S = \{u_1, \ldots, u_n \}$ and apply \Lem{geom} to the new set $S \backslash S_1$, where $S_1 = \{\tilde{u}_1\}$. Repeat this procedure $k$ times and let $S_k$ denote the set of $k$ points, $\{\tilde{u}_1, \ldots, \tilde{u}_k\}$, that satisfy inequality (\ref{eq:property}) for the sets $S, S \backslash S_1, \ldots, S \backslash S_{k-1}$ respectively. \Lem{geom} guarantees the existence of a point $u \in S \backslash S_k$ and viewing direction vector $v$ such that
\begin{eqnarray}
\label{eq:bound1}
\frac{1}{n-k}\sum_{u_i \in S \backslash S_{k} \atop u_i \neq u}^{}{ \left\langle \frac{u_i - u}{\lVert u_i - u \rVert}, v \right\rangle} & \geq & \frac{1}{2}.
\end{eqnarray}
The Cauchy-Bunyakovsky-Schwarz inequality tells us that 
\begin{eqnarray}
\label{eq:bound2}
\left\langle \frac{u_i - u}{\lVert u_i - u \rVert}, v \right\rangle & \geq & -1,
\end{eqnarray}
for all $u_i \in S_k$. Inequalities (\ref{eq:bound1}) and (\ref{eq:bound2}) together imply that
\begin{eqnarray}
\label{eq:bound3}
\sum_{i=1 \atop u_i \neq u}^{n}{ \left\langle \frac{u_i - u}{\lVert u_i - u \rVert}, v \right\rangle} \amp \geq \amp \frac{n-k}{2} - k 
\end{eqnarray}
Finally, for $k \leq n/6$, we see that the right hand side of (\ref{eq:bound3}) is bounded below by $n/4$ which implies the desired result.
\end{proof}

\bibliographystyle{asa}
\bibliography{ref}

\end{document}